\documentclass{article}

\usepackage{arxiv}

\usepackage[utf8]{inputenc} 
\usepackage[T1]{fontenc}    
\usepackage{hyperref}       
\usepackage{url}            
\usepackage{booktabs}       
\usepackage{amsfonts}       
\usepackage{nicefrac}       
\usepackage{microtype}      
\usepackage{lipsum}
\usepackage{times}
\usepackage{latexsym}
\usepackage{caption}
\usepackage{subcaption}
\usepackage{url}
\usepackage{times}
\usepackage{float}
\usepackage{refcount}
\usepackage{blindtext}
\usepackage{graphicx}
\usepackage{multirow}
\usepackage{enumitem}
\usepackage{url}
\usepackage{enumitem}
\usepackage{graphicx, array, blindtext}
\usepackage{soul}
\usepackage{longtable}
\usepackage{tabularx}
\usepackage{array}
\usepackage{pgfplots}
\usepackage{flushend}
\usepackage{calc,xcolor}
\usepackage[zerostyle=d]{newtxtt}
\usepackage[T1]{fontenc}

\newcommand{\tag}[1]{{\texttt{#1}}}

\title{CITE: A Corpus of Image--Text Discourse Relations}

\author{Malihe Alikhani\\
  Computer Science \\
  Rutgers University \\ 
  malihe.alikhani\\
  @rutgers.edu 
  \\\And
  Sreyasi Nag Chowdhury \\
  Max Planck Institute\\
  for Informatics\\ 
 sreyasi\\
@mpi-inf.mpg.de \\\And
  Gerard de Melo \\ 
  Computer Science \\
  Rutgers University \\ 
gerard.demelo\\
@rutgers.edu \\ 
\And
Matthew Stone\\
  Computer Science \\
  Rutgers University \\
  matthew.stone\\
  @rutgers.edu \\
  }

\begin{document}
\maketitle

\begin{abstract}

This paper presents a novel crowd-sourced resource for multimodal discourse: our resource characterizes inferences in image--text contexts in the domain of cooking recipes in the form of coherence relations. 
Like previous corpora annotating discourse structure between text arguments, such as the Penn Discourse Treebank, our new corpus aids in establishing a better understanding of natural communication and common-sense reasoning, while our findings have implications for a wide range of applications, such as understanding and generation of multimodal documents.

\end{abstract}


\section{Introduction}
``\emph{Sometimes a picture is worth the proverbial thousand words; sometimes a few well-chosen words are far more effective than a picture}'' -- \cite{feiner1991automating}.
Modeling how visual and linguistic information can jointly contribute to coherent and effective communication is a longstanding open problem with implications across cognitive science.  As \cite{feiner1991automating} already observe, it is particularly important for automating the understanding and generation of text--image presentations.\\
Theoretical models have suggested that images and text fit together into integrated presentations via \emph{coherence relations} that are analogous to those that connect text spans in discourse; see  \cite{8397019} and Section~\ref{sec:related}. 
%
%
This paper follows up this theoretical perspective through systematic corpus investigation. 
We are inspired by research on text discourse, which has led to large-scale corpora with information about discourse structure and discourse semantics. The Penn Discourse Treebank (PDTB) is one of the most well-known examples \cite{MiltsakakiPJW04,DBLP:conf/lrec/PrasadDLMRJW08}.  However, although multimodal corpora increasingly include discourse relations between linguistic and non-linguistic contributions, particularly for utterances and other events in dialogue \cite{cuayahuitl2015strategic,hunter:etal:2015},  
to date there has existed no dataset describing the coherence of text--image presentations.  In this paper, we describe the construction of an annotated corpus that fills this gap, and report initial analyses of the communicative inferences that connect text and accompanying images in this corpus.
As we describe in Section \ref{img--txt}, our approach asks annotators to identify the presence of specific inferences linking text and images, rather than to use a taxonomy of coherence relations.  This enables us to deal with the distinctive discourse contributions of photographic imagery.  We describe our data collection process in Section~\ref{data}, showing that our annotation scheme allows us to get reliable labels by crowdsourcing.  We present analyses in Section~\ref{analysis} that show that our annotation highlights a range of cases where text and images work together in distinctive and theoretically challenging ways, and discuss the implications of our work for the understanding and generation of multimodal documents.  We conclude in Section \ref{conclusion} with a number of problems for future research. 

\section{Introduction}
``\emph{Sometimes a picture is worth the proverbial thousand words; sometimes a few well-chosen words are far more effective than a picture}'' --\cite{feiner1991automating}.
Modeling how visual and linguistic information can jointly contribute to coherent and effective communication is a longstanding open problem with implications across cognitive science.  As \cite{feiner1991automating} already observe, it is particularly important for automating the understanding and generation of text--image presentations.\\
Theoretical models have suggested that images and text fit together into integrated presentations via \emph{coherence relations} that are analogous to those that connect text spans in discourse; see  \cite{8397019} and Section~\ref{sec:related}. \\
This paper follows up this theoretical perspective through systematic corpus investigation. 

We are inspired by research on text discourse, which has led to large-scale corpora with information about discourse structure and discourse semantics. The Penn Discourse Treebank (PDTB) is one of the most well-known examples \cite{MiltsakakiPJW04,DBLP:conf/lrec/PrasadDLMRJW08}. However, although multimodal corpora increasingly include discourse relations between linguistic and non-linguistic contributions, particularly for utterances and other events in dialogue \cite{cuayahuitl2015strategic,hunter:etal:2015}, to date there has existed no dataset describing the coherence of text--image presentations.  In this paper, we describe the construction of an annotated corpus that fills this gap, and report initial analyses of the communicative inferences that connect text and accompanying images in this corpus.

As we describe in Section \ref{img--txt}, our approach asks annotators to identify the presence of specific inferences linking text and images, rather than to use a taxonomy of coherence relations.  This enables us to deal with the distinctive discourse contributions of photographic imagery.  We describe our data collection process in Section~\ref{data}, showing that our annotation scheme allows us to get reliable labels by crowdsourcing.  We present analyses in Section~\ref{analysis} that show that our annotation highlights a range of cases where text and images work together in distinctive and theoretically challenging ways, and discuss the implications of our work for the understanding and generation of multimodal documents.  We conclude in Section \ref{conclusion} with a number of problems for future research. 

\vspace{3mm}
\section{Discourse Coherence and Text--Image Presentations}
\label{img--txt}
\label{sec:related}

\begin{figure*}[h!]
    \centering
    \begin{subfigure}[t]{0.5\columnwidth}
        \centering
        \includegraphics[width=4.7cm]{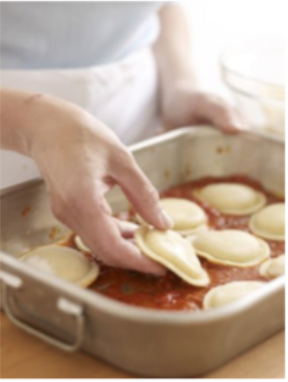}
        \caption{\textsc{Text:} Cover with a single layer of ravioli.}
    \end{subfigure}%
    ~
    \begin{subfigure}[t]{0.5\columnwidth}
        \centering
        \includegraphics[width=4.7cm]{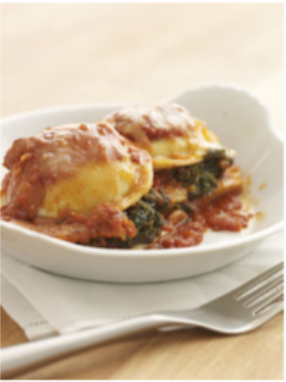}
        \caption{\textsc{Text:} Let cool 5 minutes before spooning onto individual plates.}
    \end{subfigure}
    \caption{Two steps in a recipe from \protect{\cite{DBLP:conf/emnlp/YagciogluEEI18}} illustrating diverse inferential relationships between text and accompanying imagery in instructions. The recipe is from Autodesk Inc. www.instructables.com and is contributed by www.RealSimple.com.}
    \label{fig:example}
\end{figure*}

We begin with an example to motivate our approach and clarify its relationship to previous work.
Figure~\ref{fig:example} shows two steps in an online recipe for a ravioli casserole from the RecipeQA data set \cite{DBLP:conf/emnlp/YagciogluEEI18}.
The image of Figure~\ref{fig:example}a shows a moment towards the end of carrying out the ``covering'' action of the accompanying text; that of Figure~\ref{fig:example}b shows one instance of the result of the ``spooning'' actions of the text.  
Cognitive scientists have argued that such images are much like text contributions in the way their interpretation connects to the broader discourse.
In particular, inferences analogous to those used to interpret text seem to be necessary with such images to recognize their spatio-temporal perspective \cite{cumming2017conventions}, the objects they depict \cite{abusch2012applying}, and their place in the arc of narrative progression \cite{mccloud1993understanding,cohn2013visual}.
In fact, such inferences seem to be a general feature of multimodal communication, applying also in the coherent relationships of utterance to co-speech gesture \cite{lascarides2009discourse} or the coherent relationships of elements in diagrams \cite{alikhani2018arrows,inproceedings}.

In empirical analyses of text corpora, researchers in projects such as the Penn Discourse Treebank \cite{MiltsakakiPJW04,DBLP:conf/lrec/PrasadDLMRJW08} have been successful at documenting such effects by annotating discourse structure and discourse semantics via coherence relations.
We would like to apply a similar strategy to text--image documents like that shown in Figure~\ref{fig:example}.
However, existing discourse annotation guidelines depend on the distinctive ways that coherence is signaled in text.  
In text, we find syntactic devices such as structural parallelism, semantic devices such as negation, and pragmatic elements such as discourse connectives, all of which can help annotators to recognize coherence relations in text. 
Images lack such features.
At the same time, characterizing the communicative role of imagery, particularly photographic imagery, involves a special problem: distinguishing the content that the author specifically aimed to depict from merely incidental details that happen to appear in the scene \cite{stone2015meaning}. 

Thus, rather than start from a taxonomy of discourse relations like that used in PDTB, we characterize the different kinds of inferential relationships involved in interpreting imagery separately.
\begin{itemize}
\itemsep0em
\parsep0em
\parskip0em
    \item To characterize temporal relationships between imagery and text, we ask if the image gives information about the preparation, execution or results of the accompanying step.
    \item To characterize the logical relationship of imagery to text, we ask if the image shows one of several actions described in the text, and if it depicts an action that needs to be repeated.
    \item To characterize the significance of incidental detail, we ask a range of further questions (some relevant specifically to our domain of instructions), asking about what the image depicts from the text, what it leaves out from the text, and what it adds to the text.
\end{itemize}
Our approach is designed to elicit judgments that crowd workers can provide quickly and reliably.

This approach allows us to highlight a number of common patterns that we can think of as prototypical coherence relations between images and text.
Figure \ref{fig:example}a, for example, instantiates a natural \textbf{Depiction} relation: the image shows the action described in the text in progress; the mechanics of the action are fully visible in the image, but the significant details in the imagery are all reported in the text as well.
Our approach also lets us recognize more sophisticated inferential relationships, like the fact that Figure~\ref{fig:example}b shows an \textbf{Example:Result} of the accompanying instruction.
Many of the relationships that emerge from our annotation effort involve newly-identified features of text--image presentations that deserve further investigation: particularly, the use of loosely-related imagery to provide background and motivation for a multimodal presentation as a whole, and depictions of action that seem simultaneously to give key information about the context, manner and result of an action.

\vspace{5mm}
\section{Annotation Effort}
\label{data}

Work on text has found that text genre heavily influences both the kinds of discourse relations one finds in a corpus and the way those relations are signalled \cite{webber2009genre}.
Since our focus is on developing methodology for consistent annotation, we therefore choose to work within a single genre. We selected instructional text because of its concrete, practical subject matter and because of its step-by-step organization, which makes it possible to automatically group together short segments of related text and imagery.


\paragraph{Text--Image Pairs.} We base our data collection on an existing instructional dataset, RecipeQA~\cite{DBLP:conf/emnlp/YagciogluEEI18}. This is the only publicly available large-scale dataset of multimodal instructions. It consists of multimodal recipes---textual instructions accompanied by one or more images. 

We excluded documents that either have multiple steps without images or that have multiple images per set. This was so that we could more easily study the direct relationship between an image and the associated text. 
There are 1,690 documents with this characteristic in the RecipeQA train set.  To avoid overwhelming crowd workers, we further filtered those to retain only recipes with 70 or fewer words per step, for a final count of 516 documents (2,047 image--text pairs). 

\paragraph{Protocol.}  We recruit participants through Amazon Mechanical Turk. All subjects were US citizens, agreed to a consent form approved by Rutgers's institutional review board, and were compensated at an estimated rate of USD 15 an hour.

\paragraph{Experiment Interface.} Given an image and the corresponding textual instruction from the dataset, participants were requested to answer the following 10 questions. 

For Question 1, participants were asked to highlight the relevant part of the text. For the others, we solicited True/False responses.
\begin{enumerate}
\itemsep0em
\parsep0em
\parskip0em
    \item Highlight the part of the text that is most related to the image.
    \item The image gives visual information about the step described in the text. 
    \item You need to see the image in order to be able to carry out the step properly. 
    \item The text provides specific quantities (amounts, measurements, etc.) that you would not know just by looking at the picture. 
    \item The image shows a tool used in the step but \emph{not} mentioned in the text. 
    \item The image shows how to prepare before carrying out the step. 
    \item The image shows the results of the action that is described in the text. 
    \item The image depicts an action in progress that is described in the text. 
    \item The text describes several different actions but the image only depicts one. 
    \item One would have to repeat the action shown in the image many times in order to complete this step.
\end{enumerate}

The interface is designed such that if the answer to Question 8 is \tag{True}, the subject will be prompted with Question 9 and 10. Otherwise, Question 8 is the last question in the list.\footnote{The dataset and the code for the machine learning experiments are available at https://github.com/malihealikhani/CITE}

\paragraph{Agreement.} To assess the inter-rater agreement, we determine Cohen's $\kappa$ and Fleiss's  $\kappa$ values. For Cohen's $\kappa$, we randomly selected 150 image--text pairs and assigned each to two participants, obtaining a Cohen's $\kappa$ of 0.844, which indicates almost perfect agreement. For Fleiss's $\kappa$ \cite{fleiss1973equivalence,cocos2015effectively,banerjee1999beyond}, we randomly selected 50 text--image pairs, assigned them to five subjects, and computed the average $\kappa$. We obtain a score of 0.736, which indicates substantial agreement \cite{viera2005understanding}. 

\section{Analysis}
\label{analysis}

\paragraph{Overall Statistics.}  Table~\ref{raw} shows the rates of true answers for questions Q2--Q10.
Subjects reported that in 17\% of cases the images did not give any information about the step described in the accompanying text.  Such images deserve further investigation to characterize their interpretive relationship to the document as a whole.  Our anecdotal experience is that such images sometimes provide context for the recipe, which may suggest that imagery, like real-world events \cite{hunter:etal:2015},  creates more flexible discourse structures than linguistic segments on their own.  
\begin{table}[h!]
\centering
\begin{tabularx}{0.81\linewidth}{p{9mm}|p{9mm}|p{9mm}|p{9mm}|p{9mm}|p{9mm}|p{9mm}|p{9mm}|p{9mm}|p{9mm}|p{9mm}|}
\cline{2-10}
                            & Q2    & Q3    & Q4    & Q5    & Q6    & Q7    & Q8    & Q9    & Q10   \\ \hline
\multicolumn{1}{|l|}{True}  & 0.829 & 0.058 & 0.211 & 0.131 & 0.056 & 0.491 & 0.209 & 0.289 & 0.133 \\ \hline 

\end{tabularx}
\vspace{3mm}
\caption{Rate of true answers for annotation questions Q2--Q10 across the corpus.}
\label{raw}
\end{table}

\begin{table}[h]
\centering
\begin{tabularx}{0.89\linewidth}{p{9mm}|p{9mm}|p{9mm}|p{9mm}|p{9mm}|p{9mm}|p{9mm}|p{9mm}|p{9mm}|p{9mm}|p{9mm}|p{9mm}|}
\cline{2-11}

    & Q1   & Q2**   & Q3**   & Q4**   & Q5   & Q6**   & Q7**   & Q8**  & Q9*   & Q10**  \\ \hline
\multicolumn{1}{|l|}{F1}  & 0.74 & 0.86 & 0.76 & 0.85 & 0.88 & 0.92 & 0.64 & 0.83 & 0.77 & 0.92 \\ \hline
\end{tabularx}
\vspace{4mm}

\caption{SVM classification accuracy: bag-of-words features; 80-20 train-test split; 5-fold cross validation. For the first question, this distinguishes highlighted text vs.\ its complement (excluded vs.\ included).  For the rest of the questions, this distinguishes text of true instances from text of false instances, and is different from majority class baseline $^*$ at $p < 0.04$, $t= -3.5$ and $^{**}$ at $p < 0.01$, $t>|2.49|$.}

\label{svm}
\end{table}

Subjects reported that the image was required in order to carry out the instruction only for 6\% of cases. 
This suggests that subjects construe imagery as backgrounded or peripheral to the document, much as speakers regard co-speech iconic gesture as peripheral to speech \cite{schlenker2017gestural}.  Note, by contrast, that subjects characterized 12.7\% of images as introducing a new tool: this includes many cases where the same subjects say the image is not required.  In other words, subjects' intuitions suggest that coherent imagery typically does not contribute instruction content, but rather serves as a visual signal that facilitates inferences that have to be made to carry out the instruction regardless.  Our annotated examples, where imagery is linked to specific kinds of inferences, provide materials to test this idea.

%
%
\begin{figure}[h!]
\captionsetup[subfigure]{labelformat=empty}
    \centering
    \begin{subfigure}[t]{0.99\columnwidth}
        \centering
        \includegraphics[width=4.7cm]{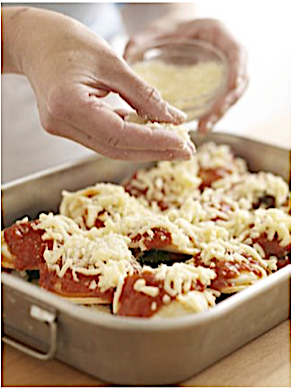}
        \caption{\textsc{Text:} Top with another layer of ravioli and the remaining sauce not all the ravioli may be needed. Sprinkle with the Parmesan.}
        \vspace{-2mm}
    \end{subfigure}%
    \caption{The image depicts both the action and the result of the action. The recipe is from Autodesk Inc.\ www.instructables.com and was contributed by www.RealSimple.com.}
    \label{fig:complex}
    \vspace{-4mm}
\end{figure}
%

\paragraph{The Complex Coherence of Imagery.}  Our annotation reveals cases where a single image does include more information than could be packaged into a single textual discourse unit (the proverbial thousand words).   In particular, such imagery participates in more complex coherence relationships than we find between text segments.  Multiple temporal relationships show this most clearly: 12\% of images that have any temporal relation have more than one.  For example, many images depict the action that is described in the text, while also showing preparations that have already been made by displaying the scene in which the action is performed.  Figure \ref{fig:complex} depicts the action and the result of the action. It also shows how to prepare before carrying out the action. Other images show an action in progress but nearing completion and thereby depict the result. For instance, the image that accompanies ``mix well until blended'' can show both late-stage mixing and the blended result. Looking at a few such cases closely, the circumstances and composition of the photos seem staged to invite such overlapping inferences.

Such cases testify to the richness of multimodal discourse, and help to justify our research methodology. The True/False questions characterize the relevant features of interpretation without necessarily mapping to single discourse relations.  For instance, Q4 and Q5 indicate inferences in line with an \textbf{Elaboration} relation; Q9 and Q10 indicate inferences in line with an \textbf{Exemplification} relation, as information presented in images show just one case of a generalization presented in accompanying text.  However, our data shows that these inferences can be combined in productive ways, in keeping with the potentially complex relevant content of images.
\begin{table*}[h]
\centering
\begin{tabularx}{0.5\linewidth}{|X|X|X|X|}
\hline
\multicolumn{4}{|>{\centering\arraybackslash}m{60mm}|}{Q4. Text has quantities not in image}                 \\ \hline
\multicolumn{2}{|c|}{True} & \multicolumn{2}{c|}{False} \\ \hline
1            & -4.1     & add        & -4.5      \\ \hline
cup          & -4.4     & place      & -4.9      \\ \hline
minutes      & -4.7     & put        & -5.0     \\ \hline
2            &  -4.7      & make       & -5.1      \\ \hline
1/2          & -4.9     & mix        & -5.1  \\\hline
\multicolumn{4}{c}{}  \\ \hline
\multicolumn{4}{|>{\centering\arraybackslash}m{60mm}|}{Q8. Image depicts action in progress}             \\ \hline
\multicolumn{2}{|c|}{True} & \multicolumn{2}{c|}{False} \\ \hline
add          &   -5.0       & 1         & -4.6     \\ \hline
mix       & -5.2     & cup           & -4.7     \\ \hline
place        & -5.3     & minutes         & -4.9     \\ \hline
bread        & -5.5     & 160     & -5.1     \\ \hline
make         &  -5.6        & put         & -5.2     \\ \hline
\end{tabularx}
\caption{Top five features of Multimodal Naive Bayes classifier for two classification problems and their corresponding log--probability estimates.}
\label{features}
\end{table*}
%
%
\paragraph{Information across modalities.}  We carried out machine learning experiments to assess what information images provide and what textual cues can guide image interpretation.  We use SVM classifiers for performance, and Multinomial Naive Bayes classifiers to explain classifier decision making, both with bag-of-words features.  
\begin{table}[]
\centering
\begin{tabular}{|l|l|l|}
\hline
\multicolumn{3}{|>{\centering\arraybackslash}m{80mm}|}{Q1. Information in text}  \\ \hline
1 & do it clearly          & on which          \\ \hline
2 & let cool for           & favorite toppings \\ \hline
3 & recipe with directions & after an          \\ \hline
4 & how slowly the         & lightly season    \\ \hline
5 & 7 minutes on           & the 2             \\ \hline
\end{tabular}

\vspace{10pt}

\begin{tabular}{|l|l|l|}
\hline
\multicolumn{3}{|>{\centering\arraybackslash}m{80mm}|}{Q1. Information in images  }  \\ \hline
1 & added a beautiful          & cover with    \\ \hline
2 & put as much           & scrapping the        \\ \hline
3 & skin off of            & finally fold        \\ \hline
4 & cut side toward        & after an            \\ \hline
5 & blend and blend        & add a               \\ \hline
\end{tabular}
\vspace{5pt}
\caption{Top five bigram and trigram features of NBSVM for the first question. The highlighted text that is most relevant to the image describes depicted actions, while the complement descriptions describe quantities or modifications of the actions that are described in the highlighted segments.}

\label{table:bigram-trigram}
\end{table}

Table~\ref{svm} reports the F1 measure for instance classification with SVMs (with 5-fold cross validation).  In many cases, machine learning is able to find cues that reliably help guess inferential patterns.  
Table~\ref{features} looks at two effective Naive Bayes classifiers, for Q4 (text has quantities) and Q8 (image depicts action in progress).  It shows the features most correlated with the classification decision and their log probability estimates.  For Q4, not surprisingly, numbers and units are positive instances.  

More interestingly, verbs of movement and combination are negative instances, perhaps because such steps normally involve material that has already been measured.  For Q8, a range of physical action verbs are associated with actions in progress; negative features correlate with steps involved in actions that don't require ongoing attention (e.g., baking).
Table~\ref{table:bigram-trigram} reports top SVM with NB (NBSVM) \cite{wang2012baselines} features for Q1 that asks subjects to highlight the part of the text that is most related to the image. Action verbs are part of highlighted text, whereas adverbs and quantitative information that cannot be easily depicted in images are part of the remaining segments of the text.
Such correlations set a direction for designing or learning strategies to select when to include imagery.



\section{Conclusions}
\label{conclusion}
In this paper, we have presented the first dataset describing discourse relations across text and imagery.  This data affords theoretical insights into the connection between images and instructional text, and can be used to train classifiers to support automated discourse analysis. Another important contribution of this study is that it presents a discourse annotation scheme for cross-modal data, and establishes that annotations for this scheme can be procured from non-expert contributors via crowd-sourcing.

Our paper sets the agenda for a range of future research.   One obvious example is to extend the approach to other genres of communication with other coherence relations, such as the distinctive coherence of images and caption text \cite{alikhani:sivl19}.  Another is to link coherence relations to the structure of multimodal discourse.  For example, our methods have not yet addressed whether image--text relations have the same kinds of subordinating or coordinating roles that comparable relations have in structuring text discourse \cite{asher2003logics}. 
Ultimately, of course, we hope to leverage such corpora to build and apply better models of multimodal communication. 

\section{Acknowledgement}
The research presented here is supported by NSF Award IIS-1526723 and  through a fellowship from the Rutgers Discovery Informatics Institute. Thanks to Gabriel Greenberg, Hristiyan Kourtev and the anonymous reviewers for helpful comments. We would also like to thank the Mechanical Turk annotators for their contributions.

\bibliographystyle{unsrt}  

\bibliography{references}





\end{document}